\documentclass{article}

\usepackage[preprint]{neurips_2023}

\usepackage[utf8]{inputenc} 
\usepackage[T1]{fontenc}    
\usepackage{hyperref}       
\usepackage{url}            
\usepackage{booktabs}       
\usepackage{amsfonts}       
\usepackage{nicefrac}       
\usepackage{microtype}      
\usepackage{xcolor}         
\usepackage{amsmath}
\usepackage{amssymb}
\usepackage{graphicx}

\title{Aggressive Post-Training Compression on Extremely Large Language Models}

\author{%
    Zining Zhang \\
National University of Singapore \\
NUS Centre for Trusted Internet and Community\\
  \texttt{zzn@nus.edu.sg} \\
   \And
   Yao Chen \\
National University of Singapore \\
   \AND
   Bingsheng He \\
National University of Singapore \\
   \And
   Zhenjie Zhang \\
   Neuron Mobility Pte. Ltd.
}

\begin{document}

\maketitle

\begin{abstract}
The increasing size and complexity of Large Language Models (LLMs) pose challenges for their deployment on personal computers and mobile devices. Aggressive post-training model compression is necessary to reduce the models' size, but it often results in significant accuracy loss. To address this challenge, we propose a novel network pruning technology that utilizes over 0.7 sparsity and less than 8 bits of quantization. Our approach enables the compression of prevailing LLMs within a couple of hours while maintaining a relatively small accuracy loss. In experimental evaluations, our method demonstrates effectiveness and potential for practical deployment. By making LLMs available on domestic devices, our work can facilitate a new era of natural language processing applications with wide-ranging impacts.
\end{abstract}

\section{Introduction}
\label{sec:introduction}
The emergence of large language models (LLMs) has been a significant breakthrough in the field of natural language processing (NLP) due to their remarkable capability in enhancing daily work productivity, as exemplified by ChatGPT~\cite{chatgpt:online}. Nevertheless, the successful deployment of these models necessitates the compression of vast amounts of human knowledge into neural network models containing hundreds of billions of parameters. As a result, such LLMs can only be accommodated on a limited range of platforms, and deployments on lower-end devices, such as personal computers, require aggressive parameter compression. Despite a plethora of proposed neural network compression methods in the literature that have demonstrated promising results on relatively small models ~\cite{li2021AdaPrune,nagel2020AdaRound,hubara2021NMPrune,hubara2021AccPTQCalib,kwon2022FastPTQTrans}, applying them to LLMs demands tens of thousands of GPU hours, rendering them impractical.

Consequently, there is a need for further research to develop novel and efficient post-training compression methods that cater to the unique requirements of LLMs. Such methods should also prioritize retaining the integrity of the model's performance while reducing its size and complexity. 
The compression of deep neural networks has been a crucial area of research in recent years. However, scaling up these models to handle massive amounts of data poses significant challenges. In this context, OPTQ ~\cite{frantar2023OPTQ} represents a ground-breaking advancement, as it offers a professional and formal solution for quantizing the massive OPT-175B~\cite{zhang2022opt} or BLOOM-176B~\cite{scao2022bloom} while maintaining high accuracy. By utilizing the traditional Optimal Brain Surgeon (OBS)~\cite{hassibi1993OBS} method, OPTQ can update remaining weights while simultaneously quantizing the weight with the least information, thereby significantly reducing computational overheads. To further improve the efficiency of this process for large models, OPTQ pre-computes the inverse of Hessian matrices for a sequential weight quantization order, eliminating the need for time-consuming calculations after each weight update. Impressively, OPTQ can quantize the whole LLM with about 175 billion parameters in approximately 4 hours on a single A100 GPU, making it an excellent choice for practitioners and researchers working with large-scale neural networks.

SparseGPT~\cite{frantar2023SparseGPT} is an algorithm that aims to further increase the compression rate of large language models (LLMs) by leveraging the same concept as OPTQ but applying it for LLM pruning. The key difference between these two algorithms is that while OPTQ \textit{sequentially} quantizes all weights, SparseGPT \textit{selectively} prunes weights in a column-block-wise manner to achieve 50\% parameter reduction in the targeting LLMs without compromising accuracy. When combined with OPTQ, SparseGPT can achieve up to a remarkable 1/8 compression ratio (fp16 to int4, 50\% sparsity). 

\begin{figure}
  \includegraphics[width=.5\linewidth]{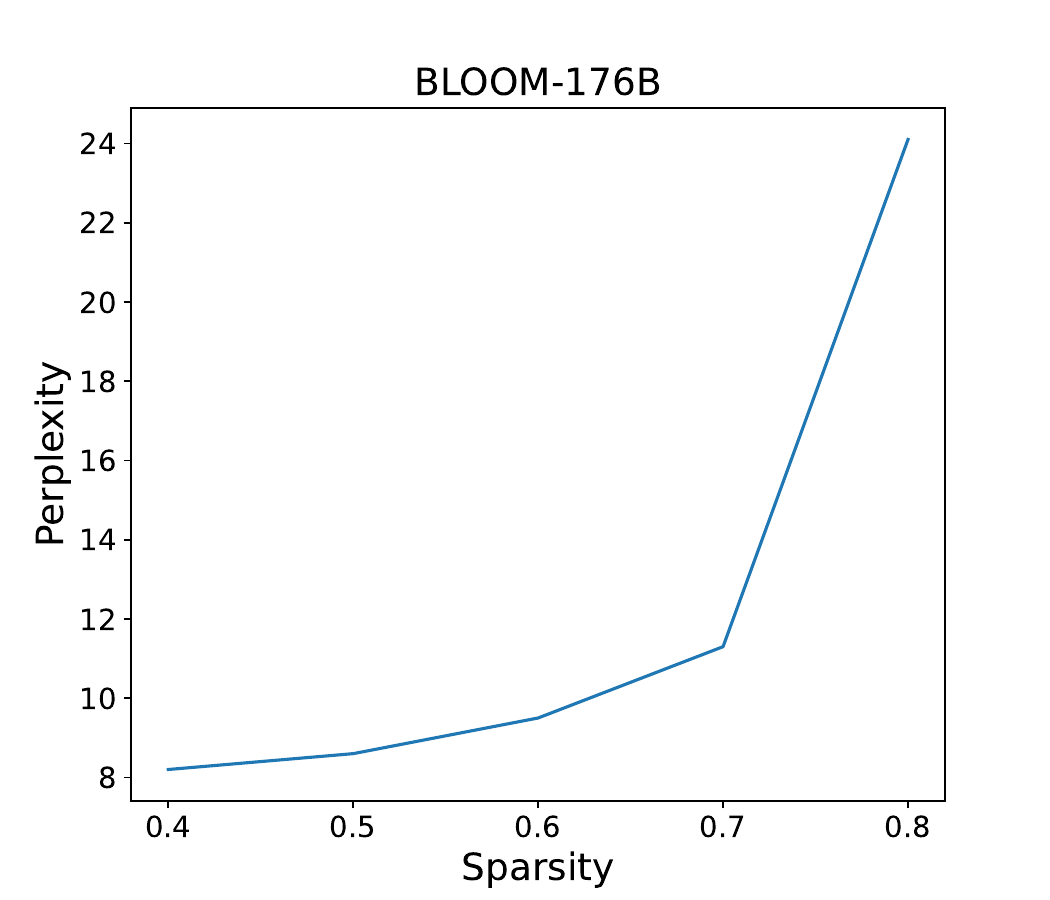}
  \centering
  \caption{The exponential relationship between sparsity and perplexity in BLOOM-176B using SparseGPT.}
  \label{fig:intro:ExpSparsPerp}
\end{figure}

While SparseGPT shows promising results in terms of accuracy and compression ratio, it presents challenges in terms of general deployability and higher sparsity. With an occupancy of over 50 GB after compressing 175 billion of parameters, the model is still not deployable on most GPU platforms. Furthermore, once the sparsity reaches 50\%, the model outputs demonstrate an exponential increase in perplexity as sparsity increases linearly, as illustrated in Figure~\ref{fig:intro:ExpSparsPerp}. 
The challenge of increasing sparsity in LLMs stems not only from the exponential increase in perplexity but also from time complexity pressures. SparseGPT's calculation of the inverse of the Hessian matrix makes an assumption that all weights are pruned in sequential order. Abandoning this assumption results in dedicated calculations of the inverse of Hessian matrices for each row vector of the weight, leading to a more than 10,000-fold increase in time complexity for massive LLMs.

To address the challenges of increasing sparsity in LLMs, we first took a step back and questioned: Why the \emph{sequentially-pruning-all} assumption in SparseGPT works in the first place? While this assumption works as expected in OPTQ, where all weights are quantized in the end, it presents a mismatch in SparseGPT due to its selective weight pruning. By resolving the mystery behind the performant \emph{sequentially-pruning-all} assumption, we developed a \emph{layer-wise sparsity scheduler} that utilizes our findings to estimate pruning errors. By controlling the sparsity distribution based on a log-level clustering of our scheduling result, we achieved higher LLM sparsity levels (>0.7) with perplexity close to dense models. The results of experiments conducted on OPT-66B and BLOOM-176B, as well as smaller LLMs, consistently outperformed the state-of-the-art. In the meantime, our method added only three hours of scheduling time for BLOOM-176B, which takes four hours for pruning.

In summary, our paper makes the following contributions:
\begin{itemize}
  \item We propose a layer-wise sparsity scheduler that utilizes pruning error estimation based on the inverse of the Hessian matrix.
  \item By employing a log-level clustering of estimated errors, we effectively control the sparsity distribution and perplexity.
  \item We provide a formal explanation for the effectiveness of the \emph{sequentially pruning all} assumption in precomputing the inverse of Hessian matrices.
  \item Our sparsity scheduler first achieves high levels of LLM sparsity (>0.7) with reasonable perplexity results.
  \item Our method is also compatible with quantization techniques that convert FP16 weights to INT4, facilitating additional compression of LLMs.
\end{itemize}

This paper is structured as follows: Section~\ref{sec:background} provides an overview of related work and background on LLM compression; Section~\ref{sec:methodology} discusses the effectiveness of the sequentially-pruning-all assumption, describes the derived novel sparsity scheduler that guides pruning, and presents pruning error estimations; Section~\ref{sec:exp} presents experimental results obtained from OPT-66B and BLOOM-176B models, as well as smaller LLMs; and Section~\ref{sec:conclusion} summarizes this paper.

\section{Background and Related Work}
\label{sec:background}

\subsection{Post-training Model Compression}
\label{ssec:bg:PostTrainingModelCompression}

The process of post-training quantization and pruning typically involves the use of a small calibration dataset to adjust quantization parameters or uncompressed weights~\cite{hubara2021AccPTQCalib,kwon2022FastPTQTrans}. Typically, compression is conducted layer-by-layer with the goal of minimizing the optimization objective:
\begin{equation}
    \label{eq:bg:layerloss}
    \operatorname{argmin}_{\widehat{W}_\ell} || \mathbf{W}_\ell \mathbf X_{\ell} - \widehat{\mathbf W}_\ell \mathbf X_\ell||^2_2
\end{equation}
Here, $\mathbf W_\ell$ and $\mathbf{\widehat{W}}_\ell$ denote the original weight and compressed weight in layer $\ell$, respectively, and $\mathbf X_\ell$ represents the input to layer $\ell$. Based on this objective, various methods have been developed. AdaRound~\cite{nagel2020AdaRound} replaces the traditional round-to-nearest quantization method by optimizing with a regularization term, while BRECQ~\cite{li2021BRECQ} and~\cite{kwon2022FastPTQTrans} utilizes Fisher information for layer quantization or pruning. AdaPrune~\cite{li2021AdaPrune} prunes model weights based on their magnitudes and compensates for errors by updating unpruned weights using gradient descent.

\subsection{Hessian-based Weight Update}
\label{ssec:bg:HessianBasedWeightUpdate}

Similar to Adaprune, Optimal Brain Compression (OBC)~\cite{frantar2022OBC} compensates for uncompressed weights during compression. However, instead of using Stochastic Gradient Descent (SGD)~\cite{ruder2016SGD}, OBC employs the Hessian matrix to directly calculate weight updates. This approach is more efficient than SGD and leads to a higher update frequency.

Let us consider the weight matrix of the current layer, denoted as $\mathbf W \in \mathbb R^{d_{row} \times d_{col}}$, where $d_{row}$ and $d_{col}$ are the dimensions of the weight matrix, and $\mathbf X \in \mathbb{R}^{d_{col} \times N}$ is the input with $N$ calibration samples. To simplify notation, we omit the subscript $\ell$. We apply a row-wise pruning error modified from Equation~\ref{eq:bg:layerloss}: $\sum_{i=1}^{d_{row}} || \mathbf{W}_{i,:} \mathbf X - \widehat{\mathbf W}_{i,:} \mathbf X ||^2_2$. We then calculate the Hessian matrix of the layer output over $\mathbf W_{i,:}$, which is represented as $\mathbf H = 2\mathbf{X} \mathbf X^\mathsf{T}$. 
When focusing on only one row $\mathbf w = \mathbf{W}_{i,:}$, the error for this specific row becomes $E= || \mathbf w \mathbf X - (\mathbf w+\delta\mathbf w) \mathbf X ||^2_2$.
Applying Taylor series expansion gives us:

\begin{equation}
\label{eq:bg:TaylorSeriesDeltaE}
\delta{ E}\ =(\frac{\partial{ E}}{\partial{\bf w}})^{\mathsf{T}}\cdot\delta{\bf w}+\frac{1}{2}\delta{\bf w}^{\mathsf{T}}\cdot{\bf H}\cdot\delta{\bf w}+{\cal O}(||\delta{\bf w}\,||^{3})
\end{equation}

Assuming the network is sufficiently trained (i.e. $\frac{\partial{ E}}{\partial{\bf w}}=0$) and ignoring the 3rd-order term, the weight updating objective becomes

\begin{equation}
\label{eq:bg:PruneObject}
\mathrm{min}_p\{\mathrm{min_{\delta \mathbf w}}(\frac{1}{2}\delta\mathbf{w^{\mathsf{T}}}\cdot\mathbf{H}\cdot\delta\mathbf{w}) { | \mathbf e_{p}^{\mathsf{T}}\cdot\delta \mathbf w+w_{p}=0\}}
\end{equation}

where $\mathbf e_p$ is a one-hot  vector which only sets on $p$th element.
This objective can be converted to a loss regularized by a Lagrangian multiplier:

\begin{equation}
\label{eq:bg:PruneObjectLag}
 L=\frac{1}{2}\delta{\bf w}^{\mathsf{T}}\cdot{\bf H}\cdot\delta{\bf w}+\lambda({\bf e}_{p}^{\mathsf{T}}\cdot\delta{\bf w}+w_{p})
\end{equation}

After solving for the local minima when pruning $w_p$, we have:

\begin{equation}
\label{eq:bg:PruneDeltaw}
\delta\mathbf{w}=-{\frac{w_{p}}{[\mathbf{H}^{-1}]_{p p}}}\mathbf{H}^{-1}\cdot\mathbf{e}_{p}
\end{equation}

\begin{equation}
\label{eq:bg:PruneL}
L=\frac{1}{2}\frac{w_{p}^{2}}{[{\bf H}^{-1}]_{p p}}
\end{equation}

Overall, the pruning process of OBC involves three steps: 1) greedily selecting the weight $w_p$ with the lowest loss $L$ using Equation~\ref{eq:bg:PruneL}, 2) updating the other weights using $\delta \mathbf w$ from Equation~\ref{eq:bg:PruneDeltaw}, and 3) recalculating the inverse of the Hessian matrix $\mathbf H^{-1}$ based on the pruned weights. Step 3 is particularly time-consuming, with a time complexity of $O(d_{col}^3)$. To address this challenge, OBC proposes to gradually update the inverse of the Hessian with a time complexity of $O(d_{col}^2)$ instead of recalculating the entire $\mathbf H^{-1}$ for every weight update:
\begin{equation}
    \label{eq:bg:Hupdate}
    \mathbf{H}_{-p}^{-1}=\left(\mathbf{H}^{-1}-{\frac{1}{[\mathbf{H}^{-1}]_{p p}}}\mathbf{H}_{;,p}^{-1}\mathbf{H}_{p,;}^{-1}\right)_{-p}
\end{equation}
where $\mathbf H^{-1}_{-p}$ represents the inverse of the Hessian matrix after $w_p$ is pruned.

To accelerate this method for LLMs, OPTQ avoids the use of greedy weight selection by quantizing weights in sequential order. This approximation has been demonstrated to be highly effective for LLMs. Furthermore, OPTQ precomputes each row of the updated $\mathbf H^{-1}_{p}$ based on Cholesky decomposition~\cite{benoit1924CholeskyDecomposition}. SparseGPT employs similar acceleration techniques, but instead of pruning all weights in sequential order, a fixed number of weights are pruned to achieve a given sparsity $S$, and the selection is based on loss $L$.

\subsection{Layer-adaptive Sparsity}
\label{ssec:bg:LayerAdaptiveSpars}
Several works have addressed the allocation of varying sparsities to different layers of neural networks, which is a similar problem to the one we tackle in this paper. For instance, LAMP~\cite{lee2020LAMP} utilizes magnitude-based pruning scores to guide sparsity allocation, but this approach does not align with our Hessian-based pruning method. Another relevant technique is Layer-adaptive OBS~\cite{dong2017LOBS}, which allocates sparsity levels to different layers based on Hessian-based scores. However, this method involves extensive computations and necessitates retraining the model.

\section{Methodology}
\label{sec:methodology}
In this section, we describe our approach to aggressive compression by delving into the rationale behind the success of the \emph{sequentially-pruning-all} assumption and its efficacy in a selective weight pruning process. Based on our findings, we derive an estimation of layer loss and demonstrate how to establish a layer-wise sparsity scheduler utilizing these loss estimation scores.

\subsection{Sequentially-Pruning-All Assumption}
\label{ssec:meth:SeqPruneOrder}

\begin{figure}
  \includegraphics[width=.5\linewidth]{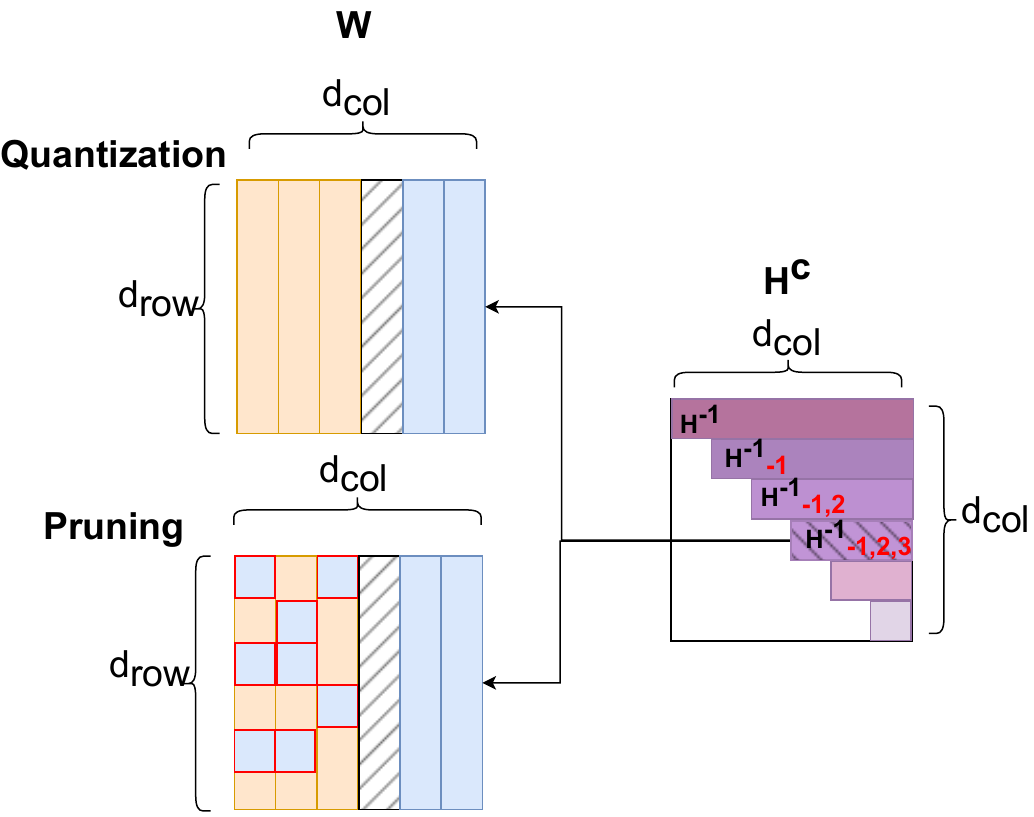}
  \centering
  \caption{Hessian-based prune vs. quantize. Sketched blocks represent the current processing column. Yellow blocks are pruned or quantized weights and blue blocks are those that haven't been compressed. Purple rows are the inverse of corresponding Hessian matrices.}
  \label{fig:meth:quantprune}
\end{figure}

In this subsection, we provide an illustration of the assumption of the \emph{sequentially-pruning-all} using Figure~\ref{fig:meth:quantprune}. The figure highlights the differences in weight updates between Hessian-based quantization and pruning. The left side of the figure displays yellow blocks representing quantized/pruned weights from previous steps and blue blocks representing yet uncompressed weights. On the right side of the figure, the upper triangular matrix $\mathbf H^c$ is the precomputed Cholesky decomposition of $\mathbf H^{-1}$, where the $i$th row is derived from $\mathbf H^{-1}_{-(1:i-1)}$ (The subscript representing the columns $(1:i-1)$ are already updated in $\mathbf H^{-1}$). The sketched blocks indicate currently processed weights or the $\mathbf H^{c}$ vector used for weight update in the current step.
The top left part of the figure shows a sequential \emph{quantization} order used in OPTQ, where all columns to the left of the current column are quantized. This aligns with the precalculated $\mathbf H^c$, as its rows assume that the $(1:i-1)$ columns are entirely quantized. However, in the context of pruning, not all $(1:i-1)$ columns are pruned, as pruning is selective based on the loss $L$ for each weight. Directly applying weight updates based on $\mathbf H^c$ results in errors, as approximately $(1-S)$ previous weights are mistakenly assumed to have been pruned when calculating $\mathbf H^c$ (depicted by blue blocks with red outlines).

Contrary to intuition, SparseGPT performs well with the \emph{sequentially-pruning-all} assumption even when setting sparsity to $S=0.5$. We explore this observation and provide an alternative perspective on the problem.

Suppose we are updating the $i$th column of the weight vector $\mathbf w$ from the previously pruned weight mask $\mathbf M^i \in \mathbf \Omega^i$, where $\mathbf \Omega^i$ represents the set of all possible masks before the $i$th column. The update is the sum of $\mathbf H^{-1,\mathbf M^j}_{j, i}$, where $j < i$ represents a previous column index, and $\mathbf H^{-1, \mathbf M^j}$ represents the inverse of the Hessian matrix after pruning by $\mathbf M^j$. 
Since the inverse Hessians are precomputed before actual pruning, we can only obtain an expectation of the Hessian inverse term for $\mathbf w_i$:

\begin{equation}
\label{eq:meth:HinvExpectation}
\sum_{j \in \mathbf M^i} {\mathbf H_{j,i}^{-1, \mathbf M^i_{:j}}} \approx \mathbb{E}_{\mathbf M^i \in \mathbf \Omega^i}\sum_{j \in \mathbf M^i} {\mathbf H_{j,i}^{-1, \mathbf M^i_{:j}}}
\end{equation}

\begin{figure}
  \includegraphics[width=.5\linewidth]{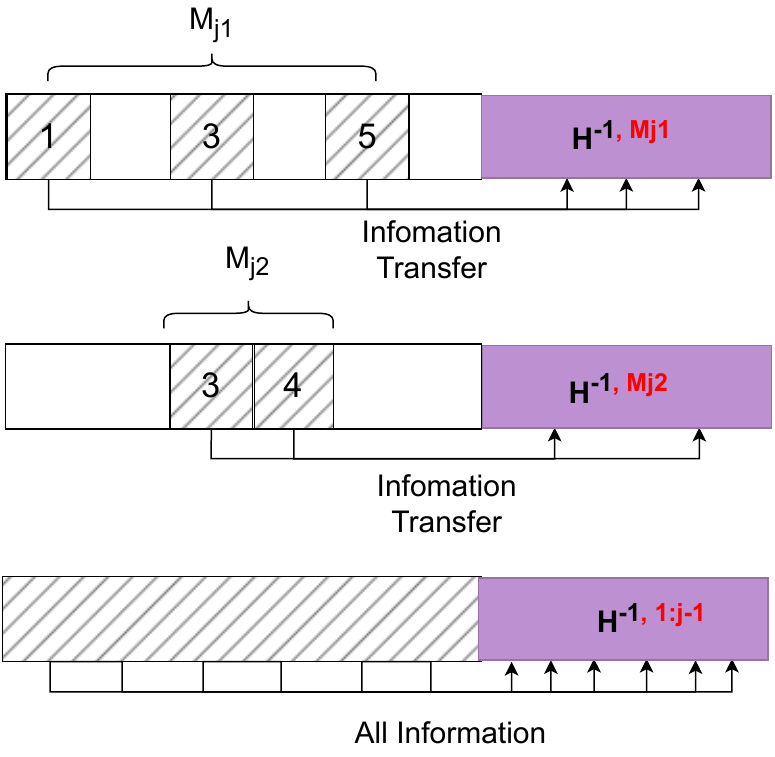}
  \centering
  \caption{The current row of $\mathbf H^c$ is the approximation of all the possible previous masks. In the top, two different masks are shown. Each mask contains the Hessian updates of previous columns, but they are reflected in $\mathbf H_{j}^{-1, {1:j-1}}$ already.}
  \label{fig:meth:HcApprox}
\end{figure}

As depicted in Figure~\ref{fig:meth:HcApprox}, $\mathbf H_{j}^{-1, {1:j-1}}$ contains all the information of the previous columns projected onto the $j$th column. Notably, after normalizing on its first element, $\mathbf H_{j}^{-1, {1:j-1}}$ represents the direction of the update that leads to the smallest pruning error \cite{hassibi1993OBS}. Therefore, we employ it as an approximation for all possible pruning subsets $M_{:j} \subseteq \{1,...,j-1\}$ since the updates to the previous columns are already reflected in $\mathbf H_{j}^{-1, {1:j-1}}$. Consequently, the equation simplifies to:
\begin{equation}
\begin{split}
    \sum_{j \in \mathbf M^i} {\mathbf H_{j,i}^{-1, \mathbf M^i_{:j}}} & \approx \mathbb{E}_{\mathbf M^i \in \mathbf \Omega^i}\sum_{j \in \mathbf M^i} \mathbf H_{j,i}^{-1, {1:j-1}} \\
    & = \sum_{j \in \{1,...,i-1\}} \text{Pr}(j\in M^i, M^i \in \Omega^j) \mathbf H_{j,i}^{-1, {1:j-1}} \\
    & \approx S \times  \sum_{j \in \{1,...,i-1\}} {\mathbf H_{j,i}^{-1, {1:j-1}}}
\end{split}
\end{equation}

The precomputed block $\mathbf H^c$ is well-suited for the task at hand, despite that it lacks a sparsity term $S$. To estimate the expectation of the weight update term $\mathbf H_{j,i}^{-1, \mathbf M^i_{:j}}$ across all $i,j$, we use $S\times\mathbf H^c$. When calculating the weight update direction using $S \times \mathbf H^c$, the sparsity $S$ cancels out:

\begin{equation}
\label{eq:meth:SCanceled}
\delta\mathbf{w}=-{\frac{w_{p}}{S[\mathbf{H}^{c}]_{p p}}}S\mathbf{H}^{c}\cdot\mathbf{e}_{p} ={\frac{w_{p}}{[\mathbf{H}^{c}]_{p p}}}\mathbf{H}^{c}\cdot\mathbf{e}_{p}
\end{equation}

Therefore, $\mathbf H^c$ succeeds under selective pruning despite being based on a \emph{sequentially-pruning-all} assumption. Intuitively, it approximates the average weight updating terms due to its ability to collect information of all the previous weights. The additional sparsity scale is canceled out during $\delta \mathbf w$ calculations.

\subsection{Sparsity Scheduler}
\label{ssec:meth:SparsityScheduler}
As discussed in Section~\ref{ssec:meth:SeqPruneOrder}, $\mathbf H^c$ provides an estimate of the weight update terms, and its rows are also involved in the loss calculation in Equation~\ref{eq:bg:PruneL}. Motivated by these observations, we propose to estimate the mean square error (MSE) for the entire layer after pruning, which can then be used to allocate different sparsity values to different layers. Specifically, we define the MSE estimation for layer $\ell$ as:

\begin{equation}
\label{eq:meth:sparssched:LayerAllScore}
L_\ell=\frac{1}{d_{row}d_{col}}\sum_{1\le i \le d_{row}}\sum_{1\le p \le d_{col}}\frac{1}{2}\frac{\mathbf W_{i,p}^{2}}{[{\bf H}^{c}]_{p p}}
\end{equation}

At first glance, this estimation may appear counterintuitive since it \emph{averages} all the scores instead of taking the \emph{sum} of estimated losses for the whole layer. This is because we allocate sparsities for different layers based on their expected loss. If we were to use the sum of the loss to represent a layer, some layers with high average losses but small numbers of elements might be allocated with high sparsity values, leading to a higher probability of pruning elements with high losses.
It is worth mentioning that we opted to compute the average of scores for each layer instead of globally ranking all scores for all weights. This decision arose due to memory constraints, as the scores for all weights easily exceed the available memory.

\begin{figure}
  \includegraphics[width=\linewidth]{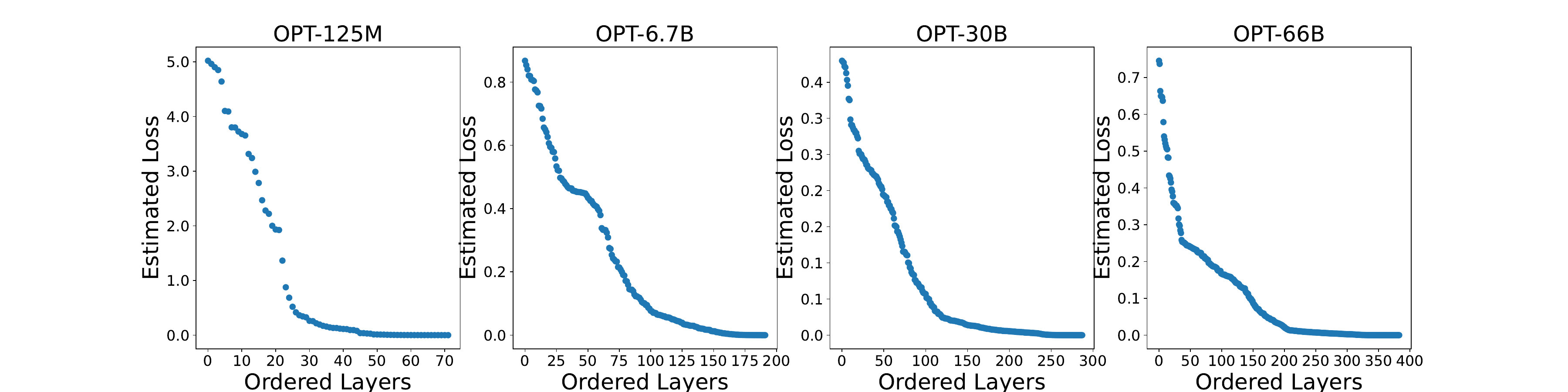}
  \centering
  \caption{Score $L_\ell$ plots for different OPT models. They all show an exponential distribution, but OPT-6.7B lacks a flat area at the low-loss region, i.e. the distribution is short-tailed.}
  \label{fig:meth:scoreplot}
\end{figure}

During our experimentation, we observed that the scores of different layers follow a exponential distribution, as illustrated in Figure~\ref{fig:meth:scoreplot}. To assign sparsities appropriately, we apply the k-means algorithm~\cite{hartigan1979kmeans} to partition the layers into $k$ groups based on $\log L_\ell$, i.e.:
\begin{equation}
    \label{eq:meth:sparssched:KMeans}
    \mathrm{argmin}_{\mu_1,\dots,\mu_k} \sum_{i=1}^{T} \sum_{j=1}^k z_{ij} (||\log L_{\ell_i} - \log \mu_j||^2)
\end{equation}
where $\mu_1,\dots,\mu_k$ denote the k cluster centroids, $z_{ij}$ is an indicator variable that equals 1 if observation $i$ belongs to cluster $j$ and 0 otherwise, $T$ is the number of layers, and $L_{\ell_i}$ is the loss estimation for layer $\ell_i$.
Considering both accuracy and compression rates, we constrain the sparsities to lie between $0.6$ and $0.8$. We ensure that the sparsity values are distributed linearly across the various clusters while ensuring that the total sparsity is at least equal to the target sparsity $S > 0.7$. The graph in Figure~\ref{fig:intro:ExpSparsPerp} reveals that there is an exponential rise in perplexity concerning sparsity that aligns with our linear increase in sparsity within the log domain of loss estimations.

\section{Experiments}
\label{sec:exp}

\subsection{Settings}
\label{ssec:exp:Settings}

To perform pruning on large language models (LLMs), we employ the open-source models BLOOM-176B~\cite{scao2022bloom} and OPT~\cite{zhang2022opt} (OPT-125M, OPT-6.7B, OPT-30B, and OPT-66B). We conduct experiments on a single Nvidia A100 GPU, fixing the sparsity setting to 0.7 during the testing of SparseGPT. Our layerwise sparsity scheduler is restricted to output only final sparsities greater than 0.7. When applying quantizations, we quantize the weights to 4 bits. 
To calibrate our method, we select 128 sequences with 2048 tokens from C4~\cite{raffel2020C4}, while we use WikiText2~\cite{merity2016Wikitext2} as the validation dataset and report perplexity as the metric. For comparison with baselines, we compare our method against SparseGPT, which is currently the state-of-the-art extremely-large LLM pruning technique, and a naive layer-wise sparsity scheduler based on the sequential order of layers.

\subsection{Sparsity vs. Perplexity}

In our experiments, we conduct sparsity scheduling on several models including OPT-125M, OPT-6.7B, OPT-30B, OPT-66B, and BLOOM-176B. The time taken by our scheduler is no longer than that of the pruning process. For instance, the sparsity scheduling for OPT-66B took 1.3 hours while the pruning process took 1.5 hours; Scheduling for BLOOM-176B took 3.4 hours and pruning took 4.5 hours.
Table~\ref{tab:exp:AccSpars} shows the perplexity scores of various language models under $\ge 0.7$ sparsity settings. Our method is to achieve no less than 70\% overall sparsity (the overall sparsity depends on the score distribution but is close to 0.7) using the sparsity scheduler introduced in Section~\ref{ssec:meth:SparsityScheduler}. We use SparseGPT with 0.7 sparsity as the baseline. Both methods were evaluated with and without quantization to 4 bits (4b). The results show that our method outperformed SparseGPT in all cases in terms of perplexity, except for OPT-6.7B. Additionally, the quantization to 4 bits did not significantly impact the performance of either method.

\begin{table*}
  \caption{Accuracy under sparsity settings $\ge 0.7$. The metric is perplexity, and the smaller perplexity means the model covers the information of the test data better. 4b represents additional quantization to INT4. 'Dense' represents original uncompressed LLMs.}
  \label{tab:exp:AccSpars}
  \centering
  \begin{tabular}{llccccc}
    \toprule
    Method            & Sparsity & OPT-125M & OPT-6.7B & OPT-30B & OPT-66B & BLOOM-176B \\
    \midrule
    Dense             & 0\%      & 27.60  & 11.53 & 9.80  &  9.32  & 8.12\\
    \midrule       
    SparseGPT         & 70\%     & 232.20 & \textbf{20.55} & 13.32 &  12.44 & 11.30\\
    SparseGPT (4b)    & 70\%     & 243.09 & \textbf{22.34} & 13.63 &  13.75 & 11.43\\
    \midrule
    Ours              & 70\%+    & \textbf{113.39} & 22.82 & \textbf{12.98} &  \textbf{11.65} & \textbf{11.02} \\
    Ours (4b)         & 70\%+    & \textbf{135.71} & 24.75 & \textbf{13.11} &  \textbf{12.84} & \textbf{11.15} \\
    \bottomrule
  \end{tabular}
\end{table*}

The cause of our method's inferior performance in OPT-6.7B is illustrated in Figure~\ref{fig:meth:scoreplot}. While most other models exhibit a significant flat region across a large proportion of layers, OPT-6.7B has a more linearly varying loss among ranked layers. This implies that allocating an aggressive range of sparsities ($[0.6, 0.8]$) is inadequate for all models. Therefore, we instead employ a narrower range of sparsities $[0.65,0.72]$, which produces the results displayed in Table~\ref{tab:exp:SmallRange6.7}. However, as of now, there is no suitable quantitative metric to determine the optimal sparsity range, and it remains a topic for future research.

\begin{table}
  \caption{The perplexity of OPT-6.7B when using a smaller sparsity range $[0.65, 0.72]$.}
  \label{tab:exp:SmallRange6.7}
  \centering
  \begin{tabular}{llc}
    \toprule
    Method            & Sparsity & OPT-6.7B \\
    \midrule
    Dense             & 0\%   & 11.53\\
    \midrule
    SparseGPT         & 70\%  & 20.55\\
    SparseGPT (4b)    & 70\%  & 22.34\\
    \midrule
    Ours              & 70\%+ & \textbf{19.73} \\
    Ours (4b)         & 70\%+ & \textbf{21.26} \\
    \bottomrule
  \end{tabular}
\end{table}

\subsection{Comparing with Naive Layer-order Sparsity Scheduler}
\label{ssec:exp:LayerOrderSparsity}

\begin{table}
  \caption{Comparison with naive layer-order sparsity scheduler.}
  \label{tab:exp:LayerOrder}
  \centering
  \begin{tabular}{llc}
    \toprule
    Method            & Sparsity & OPT-30B \\
    \midrule
    Dense             & 0\%   & 9.80\\
    \midrule
    SparseGPT         & 70\%  & 13.32\\
    Layer-order       & 70\%  & 23.67\\
    \midrule
    Ours              & 70\%+ & \textbf{12.98} \\
    \bottomrule
  \end{tabular}
\end{table}

\begin{figure}
  \includegraphics[width=0.7\linewidth]{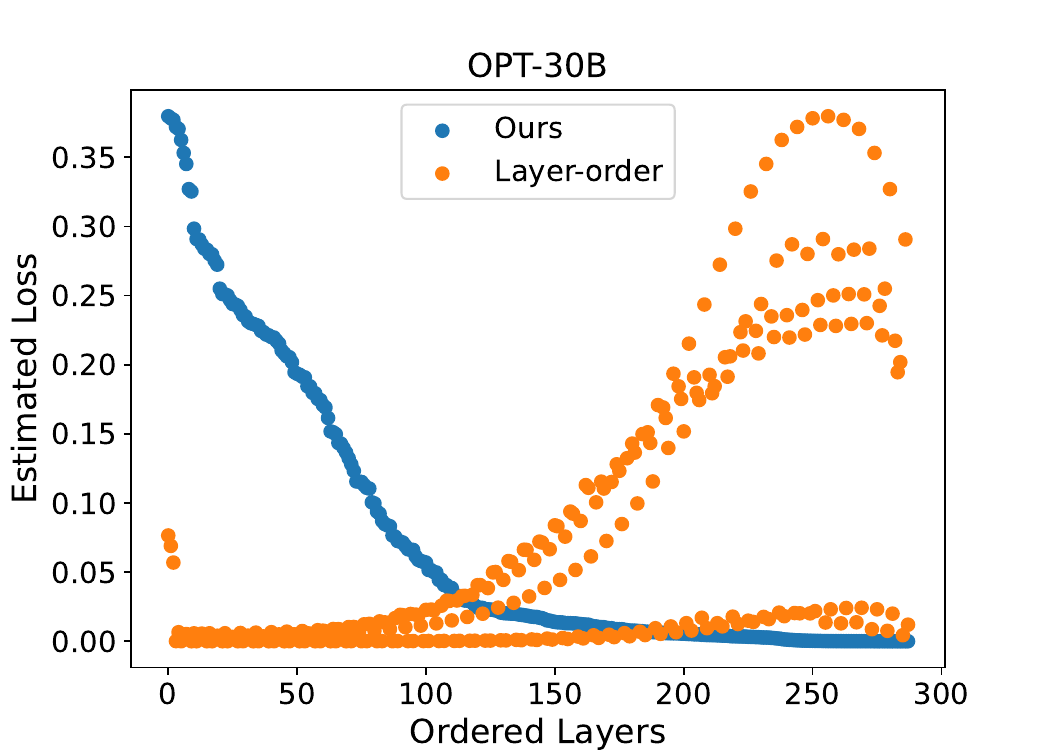}
  \centering
  \caption{Score-based ranking vs. sequential layer orders. The comparison is in our proposed loss estimation. Layer-order contains 6 different lines, since they are formed by 6 different linear modules: QKV projections, the output projection, and two forward connections.}
  \label{fig:exp:LayerOrderScheduler}
\end{figure}

As described in Section~\ref{ssec:bg:LayerAdaptiveSpars}, various techniques exist for layer-wise sparsity scheduling in neural network pruning. However, these approaches require extensive computations and are not viable for extremely large LLMs.
In addition to comparing our method with SparseGPT, which applies uniform sparsity to all layers, we also employed a naive implementation of layer-wise sparsity scheduling. The intuition was that the errors of earlier layers accumulate towards the final output, so we assigned smaller sparsities to earlier layers and larger ones to those closer to the output. Specifically, we linearly applied the same sparsity range as our scheduler ($[0.6,0.8]$) to all the layers sequentially. We performed this experiment on OPT-30B and presented the results in Table~\ref{tab:exp:LayerOrder}, where 'Layer-order' refers to the naive sparsity scheduler, which significantly harmed perplexity. To investigate the cause, we plotted our estimation scores based on the layer order in Figure~\ref{fig:exp:LayerOrderScheduler}. The graph showed that despite error accumulation across the layers, loss pruning presents an increasing behavior with layer orders. This indicates that higher layers incur more losses due to pruning. We suspect that the information contained in LLM layers becomes more critical as we approach the final output, thereby validating the effectiveness of our score-based sparsity scheduler.

\section{Conclusion}
\label{sec:conclusion}
In this paper, we proposed a novel score-based sparsity scheduler for pruning large language models (LLMs) that outperform existing techniques. Our method leverages the information provided by previous weight updates to estimate the expectation of weight updating terms under all possible pruning masks, allowing us to choose an optimal sparsity level for each layer. We demonstrated the effectiveness of our approach by comparing it with SparseGPT, the current state-of-the-art LLM pruning technique, and a naive layer-wise scheduler based on sequential order. Our experiments revealed that our method consistently outperforms SparseGPT in terms of perplexity, except for the OPT-6.7B model. We also found that a narrower sparsity range is helpful for models like OPT-6.7B which have short-tailed score distributions. Additionally, our analysis showed that higher layers contribute more losses due to pruning, indicating the importance of selecting appropriate sparsity levels for different layers. Future work could explore quantitative metrics for determining optimal sparsity ranges and investigate the relationship between sparsity and other factors, such as speed and memory usage. Overall, our score-based sparsity scheduler provides an effective and efficient solution for pruning LLMs while maintaining their perplexity.

\bibliography{neurips_2023}
\bibliographystyle{plainnat}


\end{document}